\documentclass[conference]{IEEEtran}
\usepackage[hyphens]{url}
\usepackage{graphicx}
\usepackage{cite}
\usepackage{amsmath}
\usepackage{amssymb}
\usepackage{booktabs}
\usepackage{multirow}
\usepackage{hyperref}

\hypersetup{
  hidelinks,
  colorlinks=true,
  allcolors=black,
  pdfstartview=Fit
}

\newcommand{\orcid}[1]{%
  \href{https://orcid.org/#1}{%
    \raisebox{0.1em}{\includegraphics[height=0.89em]{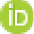}}%
  }%
}

\newcommand{\method}{UniF-MoE}

\newcommand{\ms}[2]{#1}
\newcommand{\bestms}[2]{\textbf{#1}}

\title{Share First, Route What Remains:\\
A Unified Framework for Token-Adaptive MoE Computation}
\author{
\IEEEauthorblockN{
Gongli Zhang\orcid{0009-0006-5961-6274},
Zhulin Liu\orcid{0000-0003-4145-823X}\textsuperscript{*}, and
C. L. Philip Chen\orcid{0000-0001-5451-7230}}
\IEEEauthorblockA{Guangdong Provincial Key Laboratory of Computational AI Models and Cognitive Intelligence,\\
School of Computer Science and Engineering, South China University of Technology, Guangzhou 510006, China}
\IEEEauthorblockA{Pazhou Lab, Guangzhou 510335, China}
\IEEEauthorblockA{Engineering Research Center of the Ministry of Education on Health Intelligent Perception\\
and Paralleled Digital-Human, Guangzhou 510641, China}
\IEEEauthorblockA{\textsuperscript{*}Corresponding author: Zhulin Liu.\\
Emails: csglzhang@mail.scut.edu.cn, liuzhl@scut.edu.cn, philip.chen@ieee.org}
}

\begin{document}

\maketitle

\begin{abstract}
Mixture-of-experts (MoE) models have recently moved beyond routing a fixed number of complete experts. Shared-expert designs preserve reusable knowledge, fine-grained methods vary computation within experts, and dynamic routers adapt the number of active experts. Yet these decisions are usually made independently, overlooking a basic dependency: extracting reusable computation changes both what remains and how much expert capacity the remainder needs. We study this dependency by decomposing sparsely upcycled feed-forward experts into key-value channels. Co-activated experts align at a subset of value positions; removing these positions changes expert preference; and greater shared coverage is associated with lower residual expert demand. These observations lead to one principle: share first, then route what remains. We instantiate it in \method{}, a unified framework for token-adaptive MoE computation. Each expert is partitioned into aligned blocks. A shared-demand score sets the shared block count and pathway weight, key prototypes select the shared content, and the complementary demand determines the residual expert count through cumulative routing mass. A Gram regularizer separates and normalizes router embeddings, promoting diverse routing directions, sparse expert overlap, and a simple routing geometry. Experiments on DomainBed and GLUE show that this unified design improves predictive performance over representative static and dynamic MoEs while reducing activated computation, inference latency, and memory. Code is available at \url{https://github.com/existence0420/UniF-MoE}.
\end{abstract}


\section{Introduction}

Mixture-of-experts (MoE) models increase parameter capacity without evaluating every parameter for every token \cite{firstmoebib}. A router sends each token to a small set of feed-forward networks (FFNs), or experts, so capacity can grow while execution remains sparse \cite{shazeer2017outrageously,lepikhin2021gshard}. Conventional top-$k$ routing answers one question: which experts should process this token?

This direct formulation hides two allocation decisions. It treats a complete expert as the unit of computation, even when selected experts repeat common responses, and it gives every token the same expert count, even when their semantic requirements differ. The result can be redundant computation for simple tokens and insufficient capacity for harder ones.

Recent work relaxes this formulation along three complementary directions. Shared experts capture common knowledge \cite{dai2024deepseekmoe,wu2026union}; modular, nested or slimmable experts vary computation within an expert \cite{qiu2024unlocking,jain2024mixture,tastan2026mose}; and dynamic routers vary the number of active experts \cite{huang2024harder,guo2025dynamic,park2026mass,liu2026alloc}. Other methods encourage experts to remain distinct \cite{guo2025specialization,kang2026breaking}. These advances are usually developed as separate mechanisms, but their decisions are not independent. Once reusable computation is extracted, the residual content changes, the best experts can change, and the capacity needed from them can change as well. A router that decides these quantities separately can therefore repeat shared work or allocate expert capacity for a response that has already been partly computed.

To expose this dependency, we delve into the internal structure of the FFN. An expert can be written as a sum of hidden channels, each defined by an up-projection key and a down-projection value \cite{geva2021ffn}. In a sparsely upcycled MoE \cite{komatsuzaki2023sparse}, experts start from the same pretrained FFN, so matching channel positions remain comparable across experts. This lets us test whether co-activated experts produce reusable responses at aligned positions and what happens to routing after those responses are separated.

Our analysis gives a consistent answer. Co-activated experts are neither wholly redundant nor wholly distinct: they align at a subset of value positions, and frequently co-activated pairs tend to align more. Removing these positions substantially changes the preferred experts. The number of residual experts needed to recover the original output also varies across tokens and decreases as shared coverage grows. Shared modeling, fine-grained computation, and dynamic routing are therefore three stages of one allocation problem, not three parallel knobs. The reusable response should be identified first; routing should then act on what remains.

This principle leads to \method{}, a unified framework for token-adaptive MoE computation. Each layer contains one shared expert and $K$ residual experts, all divided into aligned blocks. For each token, a shared-demand score first determines how much computation belongs to the shared pathway and how strongly that pathway contributes. Key prototypes then select which blocks provide that reusable content. Only the complementary blocks are routed onward, and their demand determines how many residual experts are activated by cumulative routing mass. A single router thus coordinates three decisions in functional order: \emph{how much to share}, \emph{what to share}, and \emph{how many experts the remainder needs}. A Gram constraint shapes its embeddings into distinct, normalized directions, encouraging diverse expert roles and sparse overlap through a simple routing geometry.

Our contributions are summarized as follows:
\begin{itemize}
\item We reveal a routing-conditioned dependency between reusable and token-specific computation inside sparsely upcycled experts. This turns shared modeling, fine-grained computation, and dynamic routing into one ordered principle: \emph{share first, then route what remains}.
\item We instantiate this principle in \method{}, which couples \emph{shared width}, \emph{shared content}, and \emph{residual expert count} through one token-dependent budget. The resulting process coordinates intra-expert and inter-expert sparsity instead of deciding them independently.
\item Across vision and language benchmarks, \method{} achieves a stronger accuracy--efficiency trade-off than representative MoEs. Ablations and routing analyses further show that its gains arise from coordinating shared reuse with residual specialization.
\end{itemize}

\section{Motivation}
\label{sec:motivation}

\subsection{Preliminaries}

Let $\mathbf{x}\in\mathbb{R}^{1\times d}$ be one input token to a sparse MoE layer with $K$ experts. The router stores one embedding per expert:
\begin{equation}
\mathbf{W}_g=[\mathbf{W}_1,\mathbf{W}_2,\ldots,\mathbf{W}_K]\in\mathbb{R}^{d\times K},
\label{eq:vanilla-router}
\end{equation}
where $\mathbf{W}_j\in\mathbb{R}^{d\times1}$ is the embedding of expert $j$. The router produces an affinity distribution and selects its $k$ largest entries:
\begin{align}
\mathbf{s}(\mathbf{x})&=\operatorname{softmax}(\mathbf{x}\mathbf{W}_g)\in\mathbb{R}^{1\times K},\label{eq:vanilla-router-score}\\
\mathcal{T}_k(\mathbf{x})&=\operatorname{TopK}(\mathbf{s}(\mathbf{x}),k),\label{eq:vanilla-topk}
\end{align}
where $\operatorname{TopK}$ returns the index set of the selected entries.

Expert $j$ is an FFN with up-projection $\mathbf{K}_j\in\mathbb{R}^{d\times H}$ and down-projection $\mathbf{V}_j\in\mathbb{R}^{H\times d}$. The intermediate width $H$ is also the number of hidden channels. Its output can be written as
\begin{equation}
E_j(\mathbf{x})=\sum_{h=1}^{H}\operatorname{GeLU}\!\left(\mathbf{x}\mathbf{K}_j[:,h]\right)\mathbf{V}_j[h,:].
\label{eq:ffn-kv}
\end{equation}
The column $\mathbf{K}_j[:,h]$ is the key of channel $h$: its inner product with $\mathbf{x}$ determines the channel activation. The row $\mathbf{V}_j[h,:]$ is the value: it determines the direction written to the output. We omit bias terms for clarity. A conventional top-$k$ MoE returns
\begin{equation}
\mathbf{y}=\sum_{j\in\mathcal{T}_k(\mathbf{x})}s_j(\mathbf{x})E_j(\mathbf{x}).
\label{eq:vanilla-moe-output}
\end{equation}
Here, $s_j(\mathbf{x})$ denotes entry $j$ of $\mathbf{s}(\mathbf{x})$. Equation~\eqref{eq:vanilla-moe-output} routes complete experts. Equation~\eqref{eq:ffn-kv}, however, shows that each expert is itself a sum of channel responses. This internal structure provides a natural place to separate shared and residual computation.

\subsection{Observations}

\paragraph{Diagnostic setup.} We analyze layer 10 of a standard top-2 MoE model trained on TerraIncognita \cite{beery2018terra}. The layer contains 6 experts with 1536 hidden channels each. All experts are copied from the same pretrained Transformer \cite{vaswani2017attention} FFN before fine-tuning. Channel index $h$ therefore refers to the same initial position across experts, which makes same-index values directly comparable. For expert pair $(i,j)$, we define
\begin{equation}
\begin{aligned}
c_{ijh}&=\frac{\mathbf{V}_i[h,:]\mathbf{V}_j[h,:]^{\top}}{\|\mathbf{V}_i[h,:]\|_2\|\mathbf{V}_j[h,:]\|_2},\\
\mathcal{S}_{ij}(\delta)&=\left\{h\in\{1,\ldots,H\}:c_{ijh}\geq\delta\right\},\\
\rho_{ij}(\delta)&=\frac{|\mathcal{S}_{ij}(\delta)|}{H}.
\end{aligned}
\label{eq:shared-set}
\end{equation}
We use $\delta=0.98$. The set $\mathcal{S}_{ij}$ contains positions where the two experts write in nearly the same direction, and $\rho_{ij}$ is the fraction of such positions. We treat this set as a weight-space estimate of the pair's reusable response. Its complement $\mathcal{R}_{ij}=\{1,\ldots,H\}\setminus\mathcal{S}_{ij}$ contains the residual positions.

\begin{figure}[t!]
\centering
\includegraphics[width=\linewidth]{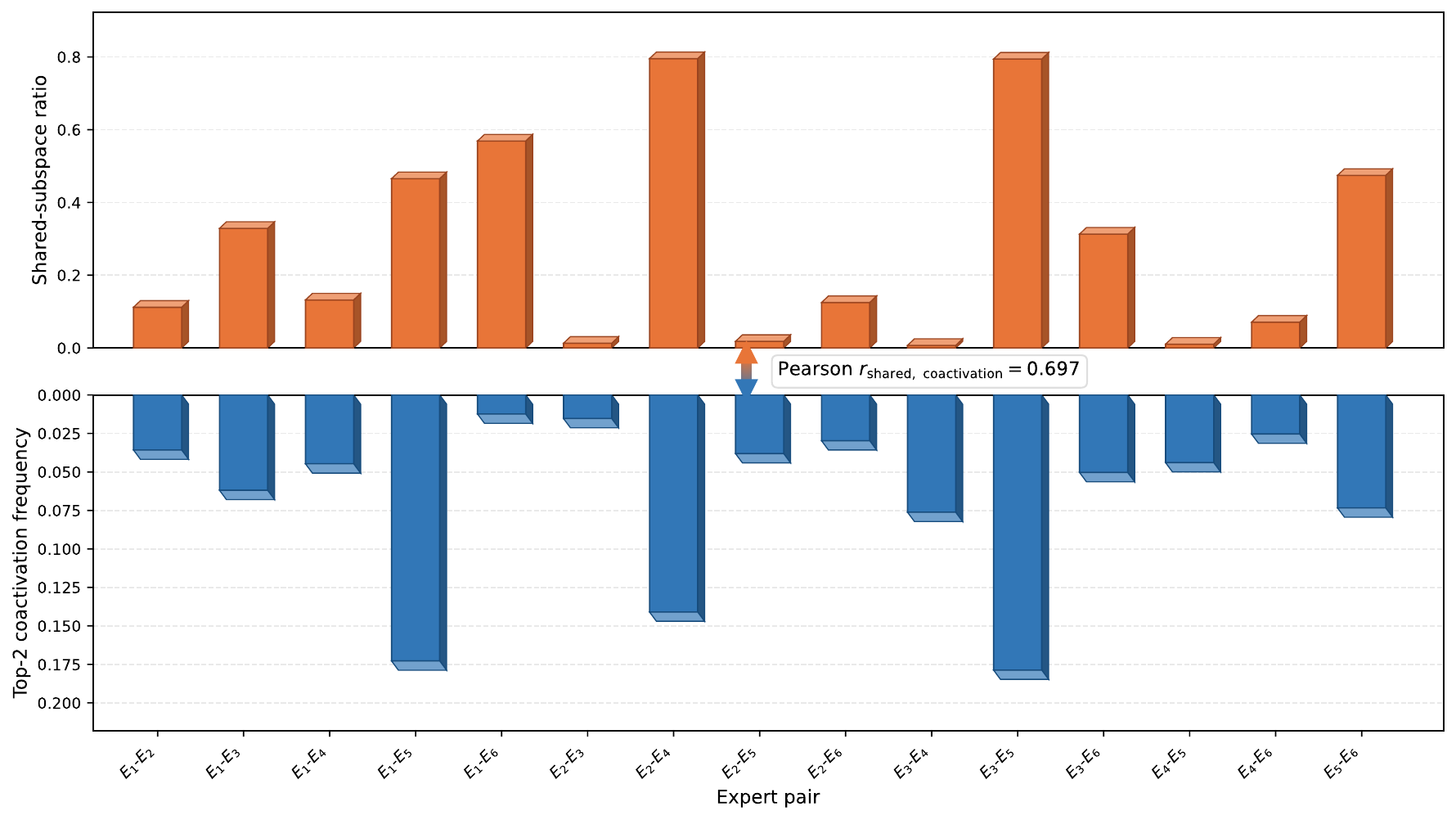}\\
(a) Shared ratio and co-activation frequency\\
\begin{tabular}{@{}c@{\hspace{0.0\linewidth}}c@{}}
\includegraphics[width=0.4\linewidth]{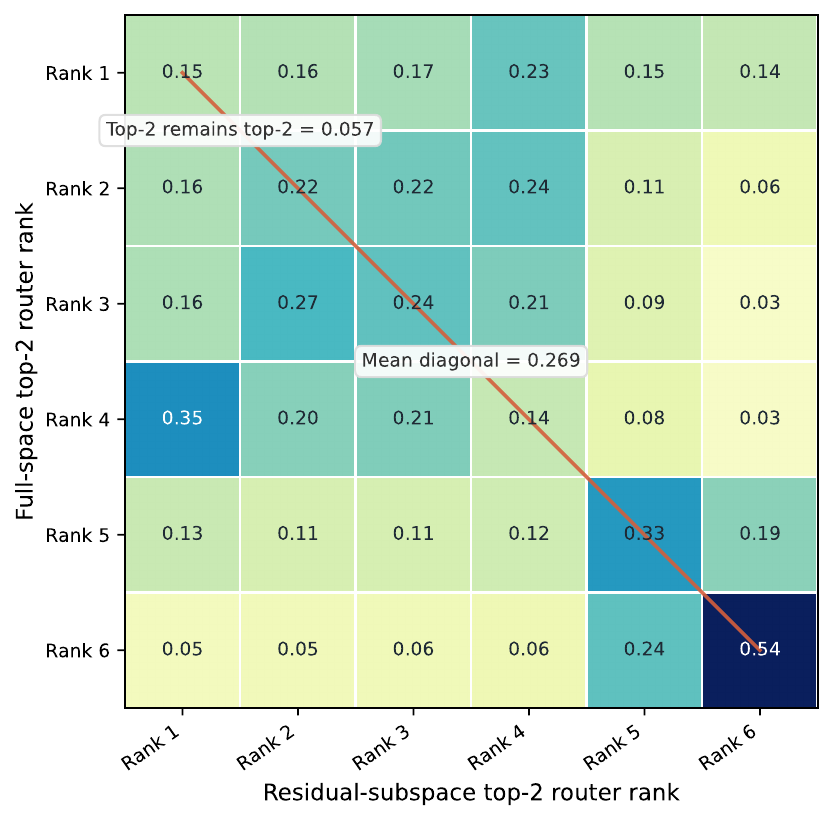} &
\includegraphics[width=0.57\linewidth]{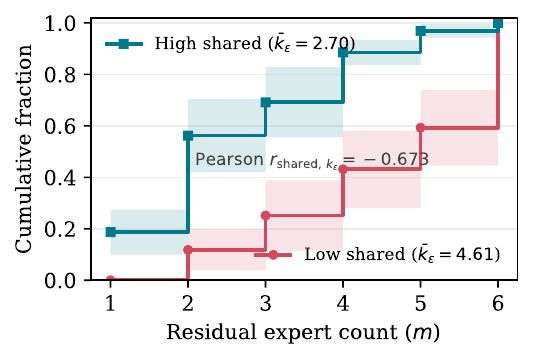} \\
(b) Router rank transition & (c) Functional-demand CDF
\end{tabular}
\caption{Diagnosing the MoE module in Transformer layer 10 trained on TerraIncognita. (a) Value alignment and co-activation for each expert pair. (b) Expert-rank transitions after the aligned positions are extracted. (c) Pair-balanced cumulative distribution function (CDF) $F_{\varepsilon}(m)=\Pr[k_{\varepsilon}(\mathbf{x})\leq m]$ at $\varepsilon=0.05$ for the five pairs with the lowest and highest shared ratios. Shading shows the standard error across pairs.}
\label{fig:observations}
\end{figure}

\paragraph{Observation 1: co-activation identifies reusable responses.} Figure~\ref{fig:observations}(a) compares $\rho_{ij}$ with the fraction of tokens routed to pair $(i,j)$. Their Pearson correlation is 0.697. Frequently co-activated pairs align at up to roughly 80\% of their value positions, while some rarely selected pairs align at less than 2\%. The aligned positions are therefore not spread uniformly across expert pairs. They concentrate among the pairs that the router most often asks to work together, which makes them a useful target for shared computation.

\paragraph{Observation 2: residual computation needs a new routing decision.} Similarity alone does not show that the aligned positions should be separated from routing. We therefore freeze the experts, extract $\mathcal{S}_{ij}$ for each token's original top-2 pair, and train a new top-2 router on the residual positions. Figure~\ref{fig:observations}(b) compares the two expert rankings. Only 5.7\% of tokens keep the same top-2 set, and the mean diagonal mass is 26.9\%. The original router scores the complete response, not the residual alone. Once the shared response is handled, different experts often become better choices.

\paragraph{Observation 3: residual demand varies across tokens.} We next ask how many residual experts are needed after the shared response has been separated. For the active pair $(i,j)$, let $\widetilde{s}_e(\mathbf{x})=s_e(\mathbf{x})/(s_i(\mathbf{x})+s_j(\mathbf{x}))$ be the normalized original gate weight. We split each expert output into $E_e^{\mathcal{S}_{ij}}$ and $E_e^{\mathcal{R}_{ij}}$, which retain the shared and residual positions, respectively. Merging the shared responses once gives
\begin{align}
\mathbf{y}_{ij}^{\mathrm{orig}}(\mathbf{x})&=\sum_{e\in\{i,j\}}\widetilde{s}_e(\mathbf{x})E_e(\mathbf{x}),\label{eq:diagnostic-original}\\
\overline{\mathbf{S}}_{ij}(\mathbf{x})&=\sum_{e\in\{i,j\}}\widetilde{s}_e(\mathbf{x})E_e^{\mathcal{S}_{ij}}(\mathbf{x}),\label{eq:diagnostic-shared}\\
\mathbf{r}_{ij}(\mathbf{x})&=\mathbf{y}_{ij}^{\mathrm{orig}}(\mathbf{x})-\overline{\mathbf{S}}_{ij}(\mathbf{x}).\label{eq:diagnostic-target}
\end{align}
The residual target is $\mathbf{r}_{ij}(\mathbf{x})$. Let $q_1(\mathbf{x}),\ldots,q_K(\mathbf{x})$ be the expert order produced by the residual router. For a prefix of length $m\in\{1,\ldots,K\}$, we stack its residual outputs:
\begin{equation}
\mathbf{R}_m(\mathbf{x})=
\begin{bmatrix}
E_{q_1(\mathbf{x})}^{\mathcal{R}_{ij}}(\mathbf{x})\\
\vdots\\
E_{q_m(\mathbf{x})}^{\mathcal{R}_{ij}}(\mathbf{x})
\end{bmatrix}
\in\mathbb{R}^{m\times d}.
\label{eq:diagnostic-residual-basis}
\end{equation}
We then use ridge-stabilized least squares to measure the best reconstruction available in the row span of $\mathbf{R}_m(\mathbf{x})$:
\begin{align}
\mathbf{a}_m^{\star}(\mathbf{x})&=\arg\min_{\mathbf{a}\in\mathbb{R}^{m}}\left\|\mathbf{a}^{\top}\mathbf{R}_m(\mathbf{x})-\mathbf{r}_{ij}(\mathbf{x})\right\|_2^2\notag\\
&\hspace{7.4em}+\gamma_m(\mathbf{x})\|\mathbf{a}\|_2^2,\label{eq:diagnostic-projection}\\
\widehat{\mathbf{y}}_m(\mathbf{x})&=\overline{\mathbf{S}}_{ij}(\mathbf{x})+(\mathbf{a}_m^{\star}(\mathbf{x}))^{\top}\mathbf{R}_m(\mathbf{x}),\label{eq:diagnostic-reconstruction}\\
e_m(\mathbf{x})&=\frac{\|\widehat{\mathbf{y}}_m(\mathbf{x})-\mathbf{y}_{ij}^{\mathrm{orig}}(\mathbf{x})\|_2}{\max\{\|\mathbf{y}_{ij}^{\mathrm{orig}}(\mathbf{x})\|_2,\eta\}},\label{eq:diagnostic-error}\\
\mathcal{A}_{\varepsilon}(\mathbf{x})&=\left\{m\in\{1,\ldots,K\}:e_m(\mathbf{x})\leq\varepsilon\right\},\label{eq:attainable-prefix}\\
k_{\varepsilon}(\mathbf{x})&=\min\left(\mathcal{A}_{\varepsilon}(\mathbf{x})\cup\{K+1\}\right).\label{eq:functional-demand}
\end{align}
We set $\gamma_m(\mathbf{x})=10^{-6}\operatorname{tr}(\mathbf{R}_m(\mathbf{x})\mathbf{R}_m(\mathbf{x})^{\top})/m$ and $\eta=10^{-12}$. $\varepsilon$ is the fidelity tolerance, and $\mathcal{A}_{\varepsilon}(\mathbf{x})$ is the set of prefixes that attain it. The sentinel $K+1$ records that no available prefix reaches the target. Thus, $k_{\varepsilon}(\mathbf{x})$ measures how quickly the ordered residual span recovers the original output. Figure~\ref{fig:observations}(c) plots its CDF with $\varepsilon$ set to 0.05. At $m=2$, the pair-balanced success rates are 56.3\% for high-shared pairs and 11.8\% for low-shared pairs; their mean $k_{\varepsilon}$ values are 2.70 and 4.61. Across pairs, shared ratio and mean demand correlate at $-0.673$ (bootstrap 95\% CI $[-0.899,-0.315]$), and the trend persists under uniform shared averaging and across all four held-out environments at $\varepsilon\in\{0.10,0.20\}$. Shared coverage is therefore related to the full distribution of residual demand.

\subsection{Design Implications}

These observations show that the key question is not whether shared modeling, fine-grained computation, and dynamic routing coexist, but in what order they act. Shared modeling must first determine how much computation is reusable. Fine-grained computation must then identify which content provides that response. Only after these choices should dynamic routing decide how many experts the remainder needs. Reversing or separating this order would route a response that has already changed. The next section implements this principle as \emph{share first, then route what remains}.

\section{Method}
\label{sec:method}

\begin{figure*}[t!]
\centering
\begin{tabular}{@{}c@{\hspace{0.01\linewidth}}c@{}}
\includegraphics[width=0.4\linewidth]{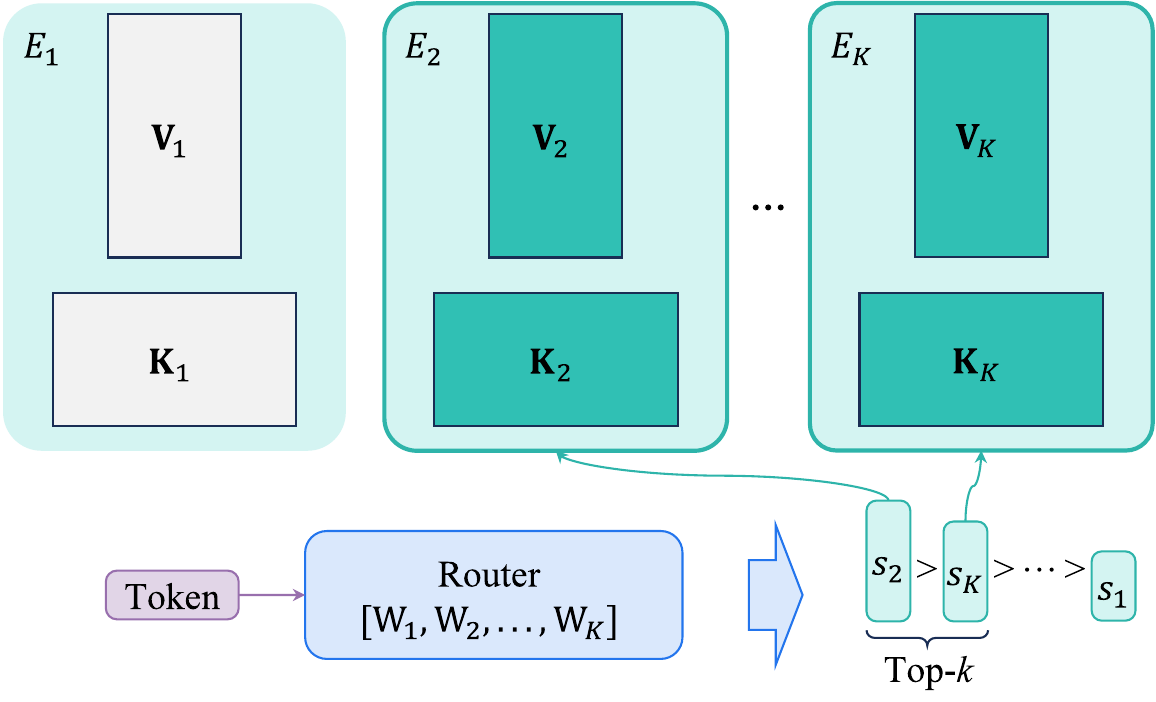} &
\includegraphics[width=0.59\linewidth]{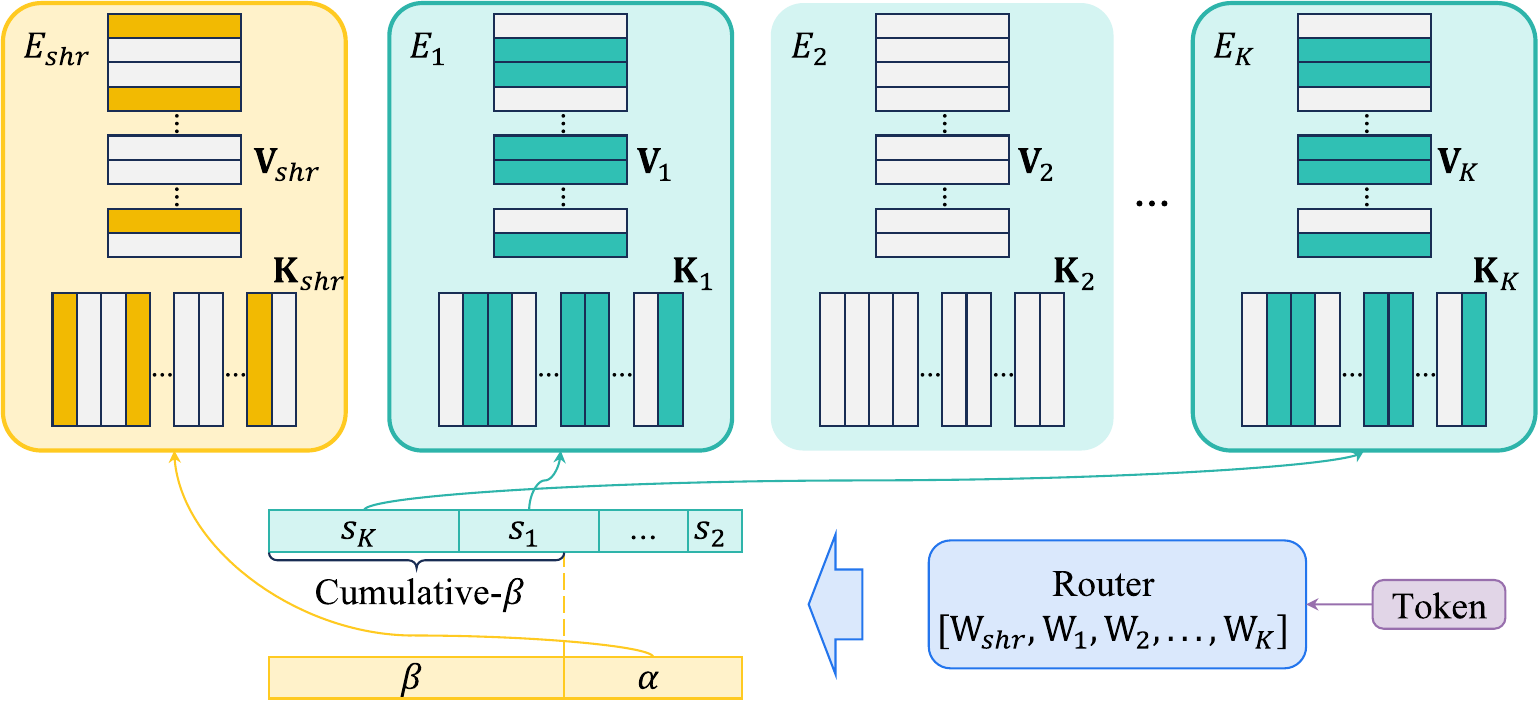} \\
(a) Conventional top-$k$ MoE & (b) \method{}
\end{tabular}
\caption{Conventional top-$k$ MoE selects complete experts in one step. \method{} first executes token-selected blocks once in a shared expert, then routes only the complementary blocks to a token-dependent number of residual experts. Colored blocks are active.}
\label{fig:architecture}
\end{figure*}

Figure~\ref{fig:architecture} compares \method{} with conventional top-$k$ MoE. Rather than attaching independent controllers for shared computation, internal selection, and expert count, \method{} makes these decisions sequentially under one token-dependent budget. 

\subsection{Blockwise Shared-Residual Partitioning}

A \method{} layer contains one shared expert $E_{\mathrm{shr}}$ and $K$ residual experts $\{E_1,\ldots,E_K\}$. Each expert has intermediate width $H$ and is divided into $B$ aligned blocks of width $M=H/B$. Block $b$ contains positions $\mathcal{H}_b=\{(b-1)M+1,\ldots,bM\}$ and produces
\begin{equation}
E_j^{(b)}(\mathbf{x})=\sum_{h=(b-1)M+1}^{bM}\operatorname{GeLU}\!\left(\mathbf{x}\mathbf{K}_j[:,h]\right)\mathbf{V}_j[h,:].
\label{eq:block-expert-output}
\end{equation}
Thus, $E_j(\mathbf{x})=\sum_{b=1}^{B}E_j^{(b)}(\mathbf{x})$. All experts are initialized from the same dense FFN, so their block boundaries are aligned at the start of training. For each token, the shared expert executes one subset of blocks, and the residual experts execute the complementary subset. Here, ``residual'' refers to the channel positions left after shared-block selection, not to the Transformer residual connection. No hidden position is assigned to both pathways for the same token.

\subsection{Token-Adaptive Shared Modeling}

\paragraph{Shared demand.} The standard router $\mathbf{W}_g$ contains the embeddings of $K$ residual experts. We add one learnable column $\mathbf{W}_{\mathrm{shr}}$ for the shared expert:
\begin{equation}
\mathbf{W}_g^{\star}=[\mathbf{W}_{\mathrm{shr}},\mathbf{W}_g]\in\mathbb{R}^{d\times(K+1)}.
\label{eq:augmented-router}
\end{equation}
This column produces the shared-demand score
\begin{equation}
\alpha(\mathbf{x})=\tau+(1-2\tau)\sigma\!\left(\mathbf{x}\mathbf{W}_{\mathrm{shr}}\right),\qquad \tau=\frac{B-1}{B^2},
\label{eq:alpha}
\end{equation}
where $\sigma(\cdot)$ is the $\operatorname{sigmoid}$ function. We interpret $\alpha(\mathbf{x})$ as the token's demand for shared computation. It controls both the number of shared blocks and the mixture weight of the shared pathway. The shared block count is
\begin{equation}
b(\mathbf{x})=\operatorname{round}(B\alpha(\mathbf{x})).
\label{eq:block-count}
\end{equation}
Because $\sigma(z)\in(0,1)$, the chosen bound gives $\tau<\alpha(\mathbf{x})<1-\tau$. For $B\geq2$, rounding therefore guarantees $b(\mathbf{x})\in\{1,\ldots,B-1\}$: every token uses at least one shared block and leaves at least one block for residual computation.

\paragraph{Shared block selection.} The score $\alpha(\mathbf{x})$ determines how many blocks to share, but not which blocks. We represent shared block $b$ by the mean of its up-projection keys:
\begin{equation}
\mathbf{\mu}_b=\frac{1}{M}\sum_{h=(b-1)M+1}^{bM}\mathbf{K}_{\mathrm{shr}}[:,h]\in\mathbb{R}^{d\times1}.
\label{eq:block-prototype}
\end{equation}
Its priority for token $\mathbf{x}$ is
\begin{equation}
u_b(\mathbf{x})=\mathbf{x}\mathbf{\mu}_b.
\label{eq:block-priority}
\end{equation}
The $b(\mathbf{x})$ blocks with the largest priorities form the shared index set. All other blocks form the residual index set:
\begin{align}
\mathcal{I}_{\mathrm{shr}}(\mathbf{x})&=\operatorname{TopK}\!\left(\{u_b(\mathbf{x})\}_{b=1}^{B},b(\mathbf{x})\right),\label{eq:shared-indices}\\
\mathcal{I}_{\mathrm{res}}(\mathbf{x})&=\{1,\ldots,B\}\setminus\mathcal{I}_{\mathrm{shr}}(\mathbf{x}).\label{eq:residual-indices}
\end{align}
Two tokens can therefore use the same shared width while choosing different content. This second decision prevents shared demand from collapsing into a fixed prefix or a scalar width rule.

\subsection{Cumulative Residual-Expert Routing}

The residual embeddings produce an affinity distribution
\begin{equation}
\mathbf{s}(\mathbf{x})=\operatorname{softmax}(\mathbf{x}\mathbf{W}_g)\in\mathbb{R}^{1\times K}.
\label{eq:residual-router-score}
\end{equation}
Let $q_1(\mathbf{x}),\ldots,q_K(\mathbf{x})$ be the expert indices sorted by decreasing affinity. We write the sorted values as $p_i(\mathbf{x})=s_{q_i(\mathbf{x})}(\mathbf{x})$:
\begin{equation}
p_1(\mathbf{x})\geq p_2(\mathbf{x})\geq\cdots\geq p_K(\mathbf{x}),\qquad p_i(\mathbf{x})=s_{q_i(\mathbf{x})}(\mathbf{x}).
\label{eq:sorted-affinity}
\end{equation}
The residual demand is the complement of the shared demand:
\begin{equation}
\beta(\mathbf{x})=1-\alpha(\mathbf{x}).
\label{eq:beta}
\end{equation}
Thus, the two pathways divide one budget instead of estimating their importance independently. We activate the smallest prefix whose cumulative affinity covers this demand:
\begin{equation}
k(\mathbf{x})=\min\left\{n\in\{1,\ldots,K\}:\sum_{i=1}^{n}p_i(\mathbf{x})\geq\beta(\mathbf{x})\right\}.
\label{eq:expert-count}
\end{equation}
Equation~\eqref{eq:expert-count} makes expert count conditional on the preceding shared allocation. The value $\beta(\mathbf{x})$ says how much computation remains, while the shape of $\mathbf{s}(\mathbf{x})$ says how concentrated that demand is across experts. A large residual can still be covered by one expert when the distribution is sharp, whereas a diffuse distribution recruits more experts.

\subsection{Shared-Residual Output Merging} 
The selected output of the shared expert and the output of residual expert $j$ are
\begin{align}
E_{\mathrm{shr}}^{\mathcal{S}}(\mathbf{x})&=\sum_{b\in\mathcal{I}_{\mathrm{shr}}(\mathbf{x})}E_{\mathrm{shr}}^{(b)}(\mathbf{x}),\label{eq:selected-shared-output}\\
E_j^{\mathcal{R}}(\mathbf{x})&=\sum_{b\in\mathcal{I}_{\mathrm{res}}(\mathbf{x})}E_j^{(b)}(\mathbf{x}).\label{eq:selected-residual-output}
\end{align}
The final layer output is
\begin{equation}
\mathbf{y}=\alpha(\mathbf{x})E_{\mathrm{shr}}^{\mathcal{S}}(\mathbf{x})+\sum_{i=1}^{k(\mathbf{x})}p_i(\mathbf{x})E_{q_i(\mathbf{x})}^{\mathcal{R}}(\mathbf{x}).
\label{eq:output}
\end{equation}
The shared output is weighted by $\alpha(\mathbf{x})$. Selected residual experts keep their affinities. Define $P_n(\mathbf{x})=\sum_{i=1}^{n}p_i(\mathbf{x})$. By the definition of the smallest sufficient prefix,
\begin{align}
\beta(\mathbf{x})&\leq P_{k(\mathbf{x})}(\mathbf{x})<\beta(\mathbf{x})+p_{k(\mathbf{x})}(\mathbf{x}),\\
1&\leq\alpha(\mathbf{x})+P_{k(\mathbf{x})}(\mathbf{x})<1+p_{k(\mathbf{x})}(\mathbf{x}).
\label{eq:mixture-mass-bound}
\end{align}
The residual routing mass is at least $\beta(\mathbf{x})$ and exceeds it by less than the affinity of the final expert. Because $\alpha(\mathbf{x})+\beta(\mathbf{x})=1$, the total coefficient mass has a controlled overshoot while preserving the router's confidence.

\subsection{Training Objective and Compute}

The shared and residual embeddings should remain distinct rather than collapse into interchangeable routes. We therefore initialize the columns of $\mathbf{W}_g^{\star}$ orthonormally and regularize their Gram matrix:
\begin{align}
\mathcal{L}_{\mathrm{div}}&=\left\|(\mathbf{W}_g^{\star})^{\top}\mathbf{W}_g^{\star}-\mathbf{I}_{K+1}\right\|_F,\label{eq:diversity-loss}\\
\mathcal{L}&=\mathcal{L}_{\mathrm{task}}+\lambda_{\mathrm{div}}\mathcal{L}_{\mathrm{div}},\label{eq:loss}
\end{align}
where $\mathbf{I}_{K+1}$ is the identity matrix. The diagonal terms keep each embedding at unit scale, yielding a simple normalized routing geometry. The off-diagonal terms separate routing directions, increasing expert diversity. Orthonormal initialization starts from the same geometry when $K+1\leq d$. Together, these effects promote diverse roles and sparse expert overlap without forcing the expert outputs themselves to be orthogonal.

Measured in blocks, token $\mathbf{x}$ activates
\begin{equation}
C_B(\mathbf{x})=b(\mathbf{x})+k(\mathbf{x})[B-b(\mathbf{x})]
\label{eq:block-compute}
\end{equation}
blocks. A conventional top-$k$ layer executes $kB$ blocks. When $k(\mathbf{x})=1$, the shared and residual blocks together equal one complete FFN. Every additional residual expert adds only the $B-b(\mathbf{x})$ blocks that remain. The ordered decomposition therefore performs reusable work once and confines repeated expert cost to the residual pathway.

\section{Experiments}
\label{sec:experiments}

\begin{table*}[t!]
\centering
\caption{Out-of-domain accuracy (\%) on DomainBed. -- and $\dagger$ denote unreported and reproduced results, respectively.}
\label{tab:domainbed-main}
\begin{tabular}{@{}lcccccccccc@{}}
\toprule
Dataset & DeiT-S/16 & GMoE & EMoE & EMoE-L & LFME & DMDA & PC-MoE & DynMoE & MASS & \textbf{\method{}} \\
\midrule
PACS & \ms{86.2}{0.1} & \ms{88.1}{0.1} & \ms{87.8}{0.2} & \ms{87.2}{0.4} & \ms{88.7}{0.2} & 88.1 & 89.4 & 88.4 & 88.7 & \bestms{89.6}{0.1} \\
VLCS & \ms{79.7}{0.0} & \ms{80.2}{0.2} & \ms{79.5}{0.4} & \ms{79.6}{0.2} & \ms{79.7}{0.1} & 79.4 & 80.0 & 80.3 & 81.1 & \bestms{81.7}{0.3} \\
OfficeHome & \ms{72.2}{0.4} & \bestms{74.2}{0.4} & \ms{73.1}{0.2} & \ms{72.5}{0.2} & \ms{73.1}{0.2} & 69.4 & 74.1 & 73.6 & 73.8 & \bestms{74.2}{0.2} \\
TerraInc. & \ms{42.0}{0.8} & \ms{48.5}{0.4} & \ms{45.9}{0.3} & \ms{46.1}{0.4} & \bestms{53.4}{0.4} & 49.5 & 49.9 & \ms{48.9}{0.5}$^\dagger$ & 47.5 & \ms{52.6}{0.3} \\
DomainNet & \ms{47.3}{0.2} & \ms{48.7}{0.2} & -- & -- & \ms{47.5}{0.0} & 45.8 & 48.5 & 48.2 & -- & \bestms{49.4}{0.1} \\
Avg & 65.5 & 67.9 & -- & -- & 68.5 & 66.4 & 68.4 & 67.9 & -- & \textbf{69.5} \\
\bottomrule
\end{tabular}
\end{table*}

\begin{table*}[t!]
\centering
\caption{In-domain GLUE task scores (\%). MoE Best is the per-task oracle upper envelope of the fixed top-$k$ variants.}
\label{tab:glue-main}
\begin{tabular}{@{}lcccccccccc@{}}
\toprule
\multirow{2}{*}{Dataset} & \multirow{2}{*}{BERT-large} & \multicolumn{6}{c}{Standard MoE} & \multirow{2}{*}{DynMoE} & \multirow{2}{*}{MASS} & \multirow{2}{*}{\textbf{\method{}}} \\
\cmidrule(lr){3-8}
& & Top-1 & Top-2 & Top-4 & Top-8 & Average & Best & & & \\
\midrule
CoLA & \ms{64.03}{0.54} & \ms{63.63}{0.20} & \ms{64.71}{1.21} & \ms{64.12}{1.42} & \ms{64.37}{1.14} & \ms{64.21}{0.45} & \ms{64.71}{1.21} & \ms{65.17}{0.26} & \ms{65.62}{1.25} & \bestms{66.83}{0.19} \\
MRPC & \ms{89.36}{0.09} & \ms{89.81}{0.30} & \ms{90.18}{1.33} & \ms{89.74}{0.40} & \ms{90.35}{0.68} & \ms{90.02}{0.29} & \ms{90.35}{0.68} & \ms{90.64}{0.26} & \ms{90.58}{0.45} & \bestms{91.57}{0.26} \\
QNLI & \ms{92.46}{0.09} & \ms{92.39}{0.21} & \ms{92.53}{0.07} & \ms{92.65}{0.09} & \ms{92.49}{0.11} & \ms{92.52}{0.11} & \ms{92.65}{0.09} & \ms{92.59}{0.08} & \ms{92.62}{0.13} & \bestms{93.10}{0.12} \\
MNLI & \ms{86.61}{0.26} & \ms{86.63}{0.17} & \ms{86.73}{0.43} & \ms{86.59}{0.16} & \ms{86.51}{0.20} & \ms{86.62}{0.09} & \ms{86.73}{0.43} & \ms{86.37}{0.13} & \ms{86.74}{0.07} & \bestms{86.84}{0.07} \\
RTE & \ms{74.37}{0.78} & \ms{74.01}{0.29} & \ms{72.32}{3.54} & \ms{75.33}{0.95} & \ms{73.53}{2.21} & \ms{73.80}{1.24} & \ms{75.33}{0.95} & \ms{73.41}{1.96} & \ms{75.33}{1.12} & \bestms{75.47}{0.91} \\
Avg & 81.37 & 81.29 & 81.29 & 81.69 & 81.45 & 81.43 & 81.95 & 81.64 & 82.19 & \textbf{82.76} \\
\bottomrule
\end{tabular}
\end{table*}

\subsection{Experimental Setup}

\paragraph{Benchmarks and implementation details.} We evaluate \method{} on vision and language tasks. For vision, the backbone is DeiT-S/16 \cite{touvron2021training} pretrained on ImageNet \cite{deng2009imagenet}. We use the train-validation selection criterion of DomainBed \cite{gulrajani2021search} and report out-of-domain accuracy on PACS \cite{li2017pacs}, VLCS \cite{fang2013vlcs}, OfficeHome \cite{venkateswara2017officehome}, TerraIncognita \cite{beery2018terra}, and DomainNet \cite{peng2019domainnet}. $K$ and $B$ are set to 6 and 8, respectively. For language, the backbone is BERT-large \cite{devlin2019bert}. We evaluate CoLA \cite{warstadt2019cola}, MRPC \cite{dolan2005mrpc}, QNLI \cite{rajpurkar2016squad}, MNLI \cite{williams2018mnli}, and RTE \cite{bentivogli2009rte} from GLUE \cite{wang2019glue}, using $K=16$ and $B=16$. Learning rate and other settings follow the corresponding baselines. Our results are averaged over three runs.

\paragraph{Comparative baselines.} Vision baselines include dense DeiT-S/16; static MoEs GMoE \cite{li2023sparse}, EMoE, and EMoE-L (EMoE-learn) \cite{qiu2024unlocking}; domain-generalization methods LFME \cite{chen2024lfme}, DMDA \cite{long2024rethinking}, and PC-MoE \cite{nguyen2025pcmoe}; and dynamic MoEs DynMoE \cite{guo2025dynamic} and MASS \cite{park2026mass}. Language baselines include dense BERT-large, fixed top-$k$ MoEs with $k\in\{1,2,4,8\}$, DynMoE, and MASS. For fixed routing, we also report the average across $k$ and the per-task best result.

\subsection{Main Results}

\paragraph{Vision tasks.} Table~\ref{tab:domainbed-main} shows that \method{} obtains the best average result on DomainBed, leading on PACS, VLCS, and DomainNet and tying the best result on OfficeHome. These four datasets mix category evidence with strong style or source variation, a setting that plays to \method{}'s strength: reusable blocks preserve transferable cues, while residual routing directs the remaining appearance-specific computation to the patches that need it. TerraIncognita is dominated by location-dependent backgrounds and camera viewpoints, where LFME's explicit domain specialization remains stronger. The overall pattern supports the value of coordinating reusable and residual computation without imposing one budget on every patch.

\paragraph{Language tasks.} Table~\ref{tab:glue-main} shows that \method{} performs best on all five GLUE tasks, spanning linguistic acceptability, paraphrase detection, and several forms of textual inference. The strongest fixed $k$ changes across tasks, yet even their per-task oracle envelope remains below \method{}. Choosing $k$ per dataset still assigns one operating point to every token. \method{} instead identifies reusable linguistic computation before adapting the residual expert count within each sequence, replacing a global expert budget with a token-level dependency.

\subsection{Token-Adaptive Computation and Cost}

\paragraph{Element-conditioned pathway composition.}

\begin{figure}[t!]
\centering
\begin{tabular}{@{}l@{\hspace{0.01\linewidth}}c@{}}
\includegraphics[height=0.44\linewidth]{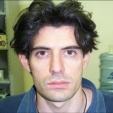} &
\includegraphics[height=0.44\linewidth]{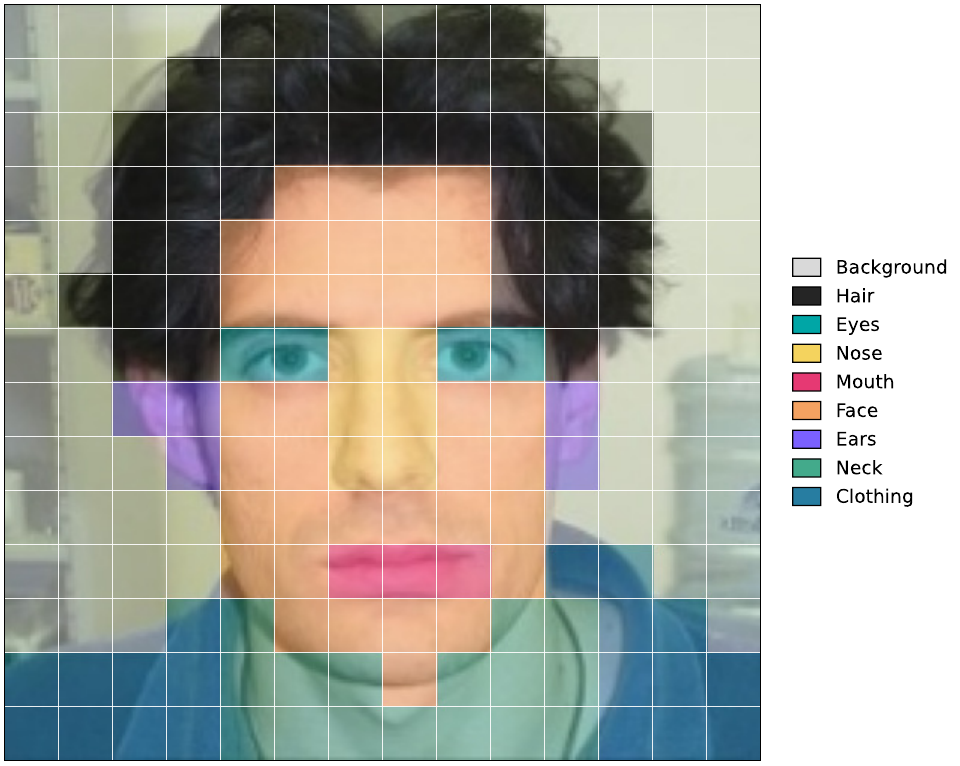} \\
(a) Original image & (b) Patch semantics \\
\end{tabular}
\includegraphics[width=\linewidth]{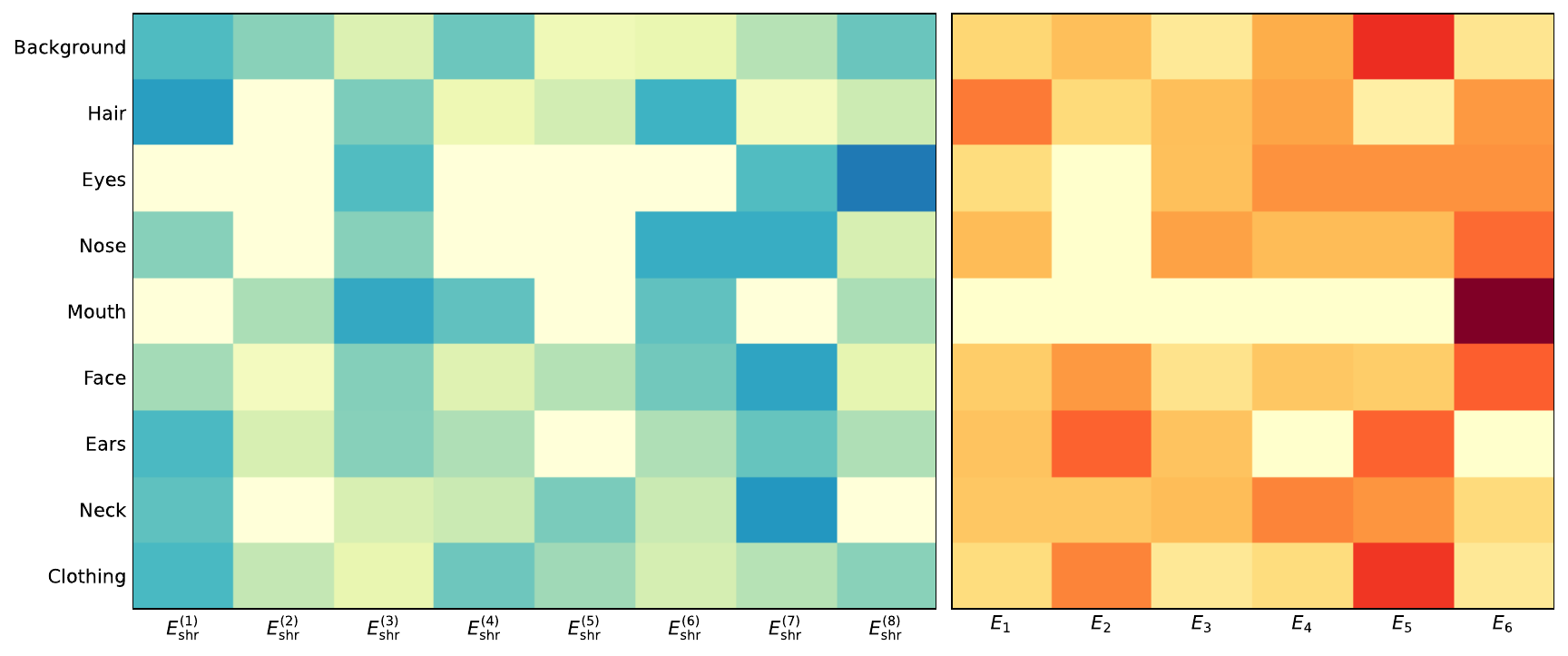} \\
(c) Shared-block and residual-expert usage \\
\caption{Element-conditioned routing for a \emph{Person} example of PACS. Darker cells indicate higher relative use within each semantic element.}
\label{fig:element-routing}
\end{figure}

Figure~\ref{fig:element-routing} groups image patches by semantic element and records their routing choices in Transformer layer 10. Shared blocks recur across several elements, whereas residual experts show more selective preferences. Patches that reuse the same shared block can still activate different residual experts, confirming the dependency at the center of our design: shared selection removes common computation, but does not predetermine who should process the remainder.

\paragraph{Patch-wise resource allocation.}

\begin{figure}[t!]
\centering
\begin{tabular}{@{}c@{\hspace{0.\linewidth}}c@{}}
\includegraphics[width=0.5\linewidth]{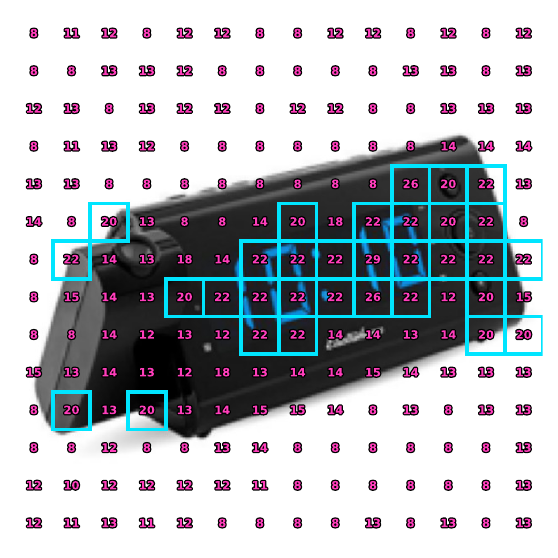} &
\includegraphics[width=0.5\linewidth]{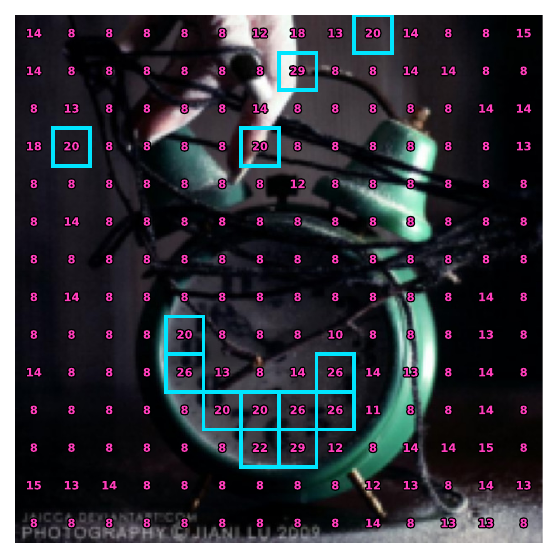} \\
(a) Product & (b) Art \\
\includegraphics[width=0.5\linewidth]{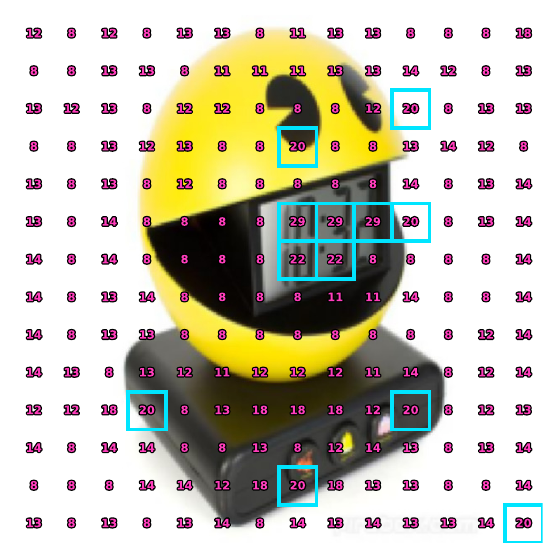} &
\includegraphics[width=0.5\linewidth]{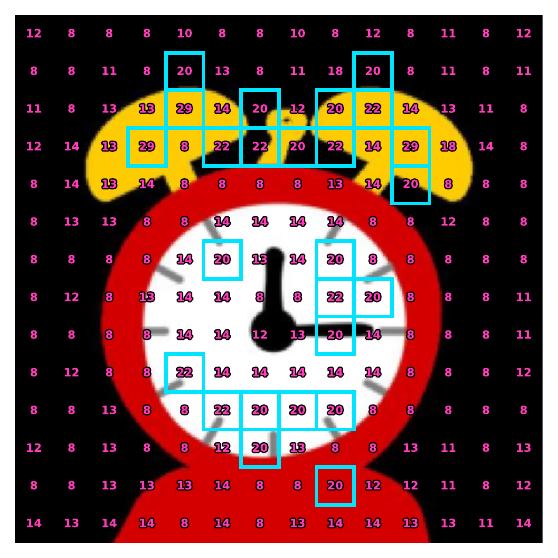} \\
(c) Real World & (d) Clipart
\end{tabular}
\caption{Patch-level block usage $C_B(\mathbf{x})$ for the \emph{Alarm Clock} class across the four OfficeHome domains. High-demand patches with $C_B(\mathbf{x})\geq20$ are outlined.}
\label{fig:patch-compute}
\end{figure}

Figure~\ref{fig:patch-compute} shows alarm clocks from different domains. High-compute patches fall on the display in Product and Real World, and on the bells, ticks, and hands in Clipart. The Art example assigns minimal computation to a watermark and several strong but irrelevant strokes. The joint budget therefore responds neither to domain nor contrast alone; it reserves residual capacity for regions that carry useful class evidence after reusable features are handled.

\paragraph{Measured cost comparison.}

Table~\ref{tab:legacy-cost} compares \method{} with representative dense, static-MoE, and dynamic-MoE designs. EMoE has the lowest activated parameter count and FLOPs among the MoEs, whereas \method{} achieves the lowest inference time and memory. Relative to top-2 GMoE, it activates 9.1\% fewer parameters and uses 16.1\% fewer FLOPs, reducing inference time and memory by 45.2\% and 52.7\%; it also undercuts DynMoE on every compute and runtime statistic. \method{} does not minimize every analytical proxy, but sharing once and repeating only residual blocks yields the strongest measured inference profile among the MoE baselines.

\begin{table}[t!]
\centering
\caption{Cost comparison on VLCS. $N$ denotes the total parameter count, and $N_A$ denotes the average activated parameter count across samples; both are reported in units of $2^{20}$ parameters. FLOPs are reported per image in units of $2^{30}$ operations. T-TPS and I-TPS denote training and inference time per step, respectively, in seconds; T-RTM and I-RTM denote the corresponding run-time memory in GiB. The best MoE values are bold.}
\label{tab:legacy-cost}
\begin{tabular}{@{}lccccc@{}}
\toprule
Statistic & DeiT-S/16 & GMoE & EMoE & DynMoE & \textbf{\method{}} \\
\midrule
$N$ & 20.66 & 32.12 & \textbf{20.70} & 36.45 & 34.19 \\
$N_A$ & 20.66 & 23.11 & \textbf{18.81} & 24.14 & 21.01 \\
FLOPs & 8.57 & 10.37 & \textbf{7.92} & 14.63 & 8.70 \\
T-TPS & 0.23 & 1.23 & \textbf{1.19} & 3.32 & 1.23 \\
I-TPS & 0.04 & 0.31 & 0.30 & 1.06 & \textbf{0.17} \\
T-RTM & 6.81 & 8.38 & \textbf{7.15} & 11.80 & 8.84 \\
I-RTM & 0.29 & 0.55 & 0.39 & 0.99 & \textbf{0.26} \\
\bottomrule
\end{tabular}
\end{table}

\subsection{Ablation Study and Hyperparameter Analysis}

\paragraph{Contribution of the three adaptive decisions.}
\begin{table}[t]
\centering
\caption{Ablating the three adaptive decisions. $\times$ indicates that the corresponding mechanism is replaced by its non-adaptive counterpart: fixed $\alpha=0.4$, top-2 residual-expert activation, or prefix-based block selection, respectively. The full block counts are 56 on DomainBed and 272 on GLUE.}
\label{tab:mechanism-ablation}
\setlength{\tabcolsep}{4.5pt}
\begin{tabular}{@{}ccc cc cc@{}}
\toprule
Adaptive-$\alpha$ & Cumulative-$\beta$ & Prototype-$\mathbf{\mu}$ & \multicolumn{2}{c}{DomainBed} & \multicolumn{2}{c}{GLUE} \\
\cmidrule(lr){4-5}\cmidrule(lr){6-7}
& & & $\bar{\mathrm{Acc}}$ & $\bar{C_B}$ & $\bar{\mathrm{Acc}}$ & $\bar{C_B}$ \\
\midrule
$\times$ & $\checkmark$ & $\checkmark$ & 68.20 & 14.17 & 81.73 & 78.33 \\
$\checkmark$ & $\times$ & $\checkmark$ & 68.90 & 13.41 & 82.07 & 66.74 \\
$\checkmark$ & $\checkmark$ & $\times$ & 68.80 & 13.06 & 81.95 & 60.14 \\
$\checkmark$ & $\checkmark$ & $\checkmark$ & \textbf{69.50} & \textbf{11.41} & \textbf{82.76} & \textbf{47.61} \\
\bottomrule
\end{tabular}
\end{table}
Table~\ref{tab:mechanism-ablation} breaks the ordered process one decision at a time. Every replacement lowers accuracy and increases computation on both benchmarks. Fixing $\alpha$ is most disruptive because an incorrect shared-residual split propagates to both later decisions. Prefix-based selection loses token-specific shared content, while top-2 residual-expert activation ignores how much demand remains. Their failures show that the gain comes from coordinating the three stages, not merely adding three forms of sparsity.

\paragraph{Effect of diversity regularization.}
\begin{figure}[t!]
\centering
\includegraphics[width=\linewidth]{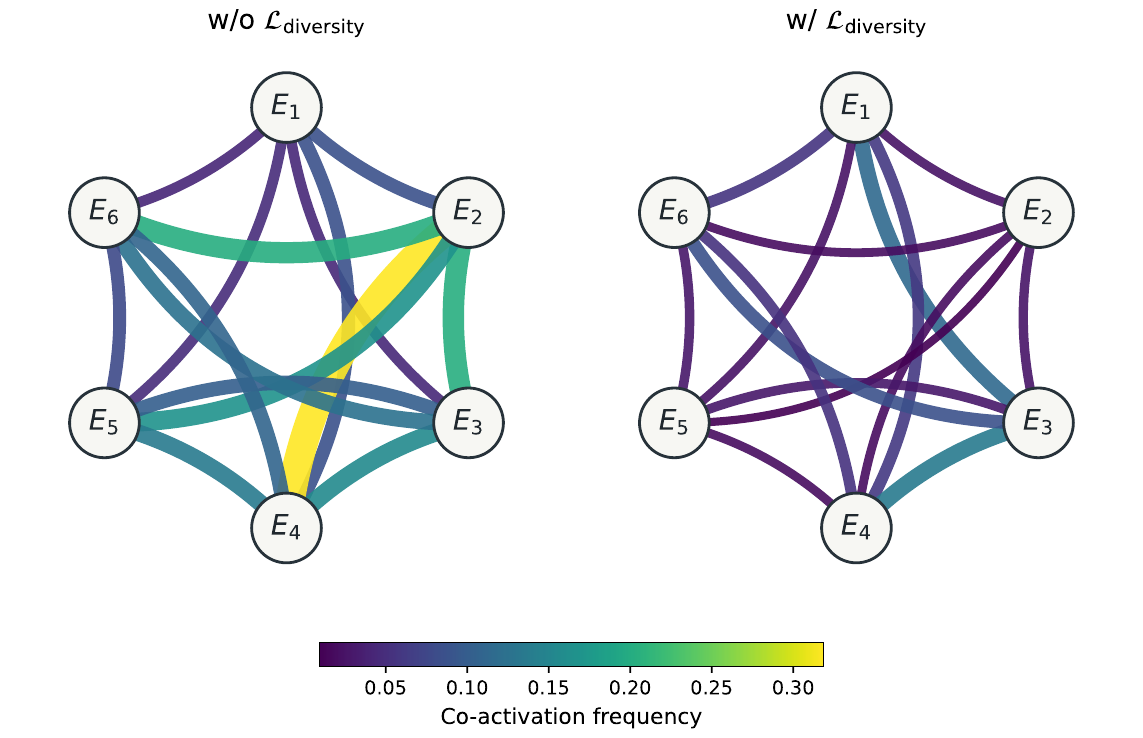} \\
(a) Residual expert co-activation \\
\includegraphics[width=0.8\linewidth]{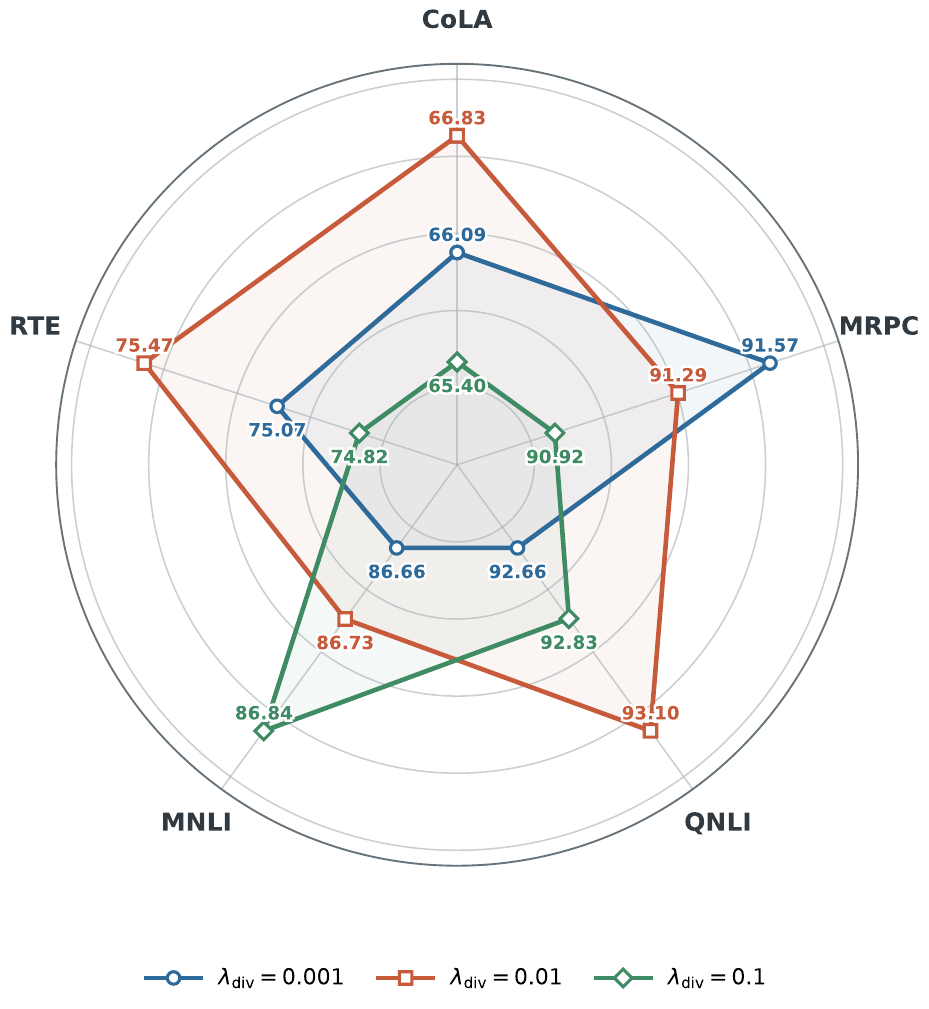} \\
(b) Diversity-loss weight
\caption{Effect of diversity regularization. (a) Residual expert co-activation on PACS without and with $\mathcal{L}_{\mathrm{div}}$. (b) GLUE results with different values of $\lambda_{\mathrm{div}}$.}
\label{fig:diversity-analysis}
\end{figure}
Figure~\ref{fig:diversity-analysis} examines the qualitative and quantitative effects of the diversity constraint. As shown in Figure~\ref{fig:diversity-analysis}(a), it reduces mean pair co-activation by 62.9\% and moves the router embeddings close to orthonormality. The graph does not become empty: useful expert cooperation remains, but routing is no longer concentrated in the same few pairs. Figure~\ref{fig:diversity-analysis}(b) further favors the intermediate setting $\lambda_{\mathrm{div}}=0.01$ on average. A weaker constraint leaves routing directions overly correlated, whereas an excessive one can overwhelm the task loss. Together, the two panels show that moderate regularization preserves distinct residual routes without prohibiting useful cooperation.

\paragraph{Shared-block granularity.}
\begin{figure}[t!]
\centering
\includegraphics[width=\linewidth]{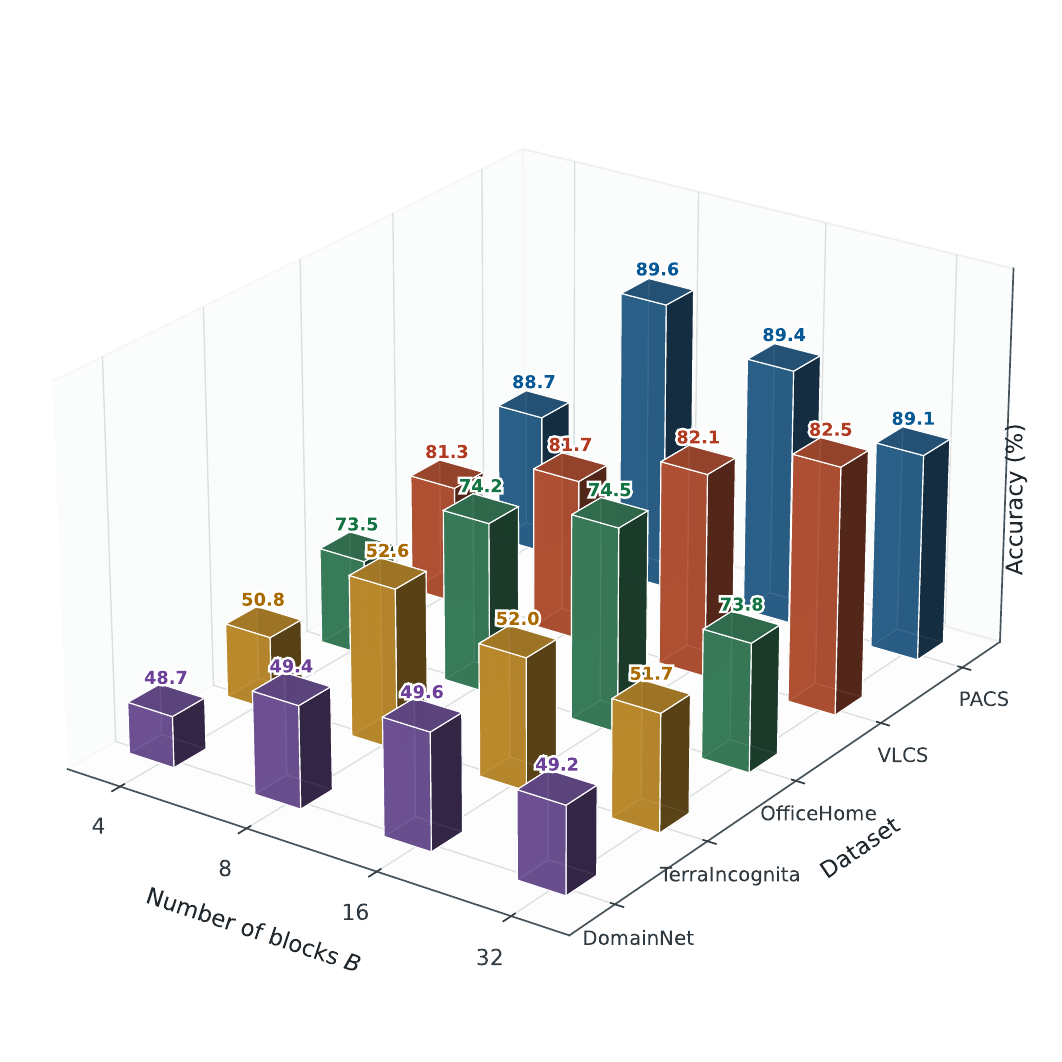}
\caption{DomainBed results with different values of $B$.}
\label{fig:shared-block-granularity}
\end{figure}
Figure~\ref{fig:shared-block-granularity} shows that $B=8$ or $16$ performs best on four of the five DomainBed datasets. With $B=4$, one selection changes a quarter of the FFN, which is too coarse to localize the shared response. With $B=32$, each prototype summarizes a narrow group of keys and cannot cover enough semantics. We use $B=8$ as a stable balance between selection granularity and complexity.

\section{Related Work}

\subsection{Shared Modeling and Expert Specialization}

Trained experts may contain overlapping computations. DeepSeekMoE finely segments experts into smaller ones and adds dedicated shared experts \cite{dai2024deepseekmoe}, while Union-of-Experts builds a virtual shared expert from routing-neuron outputs \cite{wu2026union}. A complementary line encourages expert specialization: orthogonality and variance objectives reduce overlap \cite{guo2025specialization}, and MP-MoE uses inter-expert covariance to select diverse expert sets \cite{kang2026breaking}. These methods improve reuse or specialization, but do not use token-specific shared allocation to define the computation that specialization should handle. \method{} makes that dependency explicit: the shared pathway handles reusable block responses first, and the Gram constraint preserves distinct routes for what remains.

\subsection{Dynamic Routing and Fine-grained Computation}

Dynamic MoEs vary expert count by accumulating routing confidence \cite{huang2024harder}, growing or pruning the expert pool \cite{guo2025dynamic}, combining cumulative mass with expert expansion \cite{park2026mass}, or distributing a constrained budget across layers and tokens \cite{liu2026alloc}. Fine-grained methods instead focus inside the FFN: Emergent MoE uses key centroids to expose modular structure \cite{qiu2024unlocking}, while nested and slimmable experts vary the executed width \cite{jain2024mixture,tastan2026mose}. These approaches adapt expert count or width, but generally make the two allocations independently and without conditioning either on a shared response. \method{} orders them through one shared-residual budget: the shared demand determines shared width and content, while the residual demand determines the residual expert count.

\section{Conclusion}

Shared modeling, fine-grained computation, and dynamic routing need not be separate mechanisms. By exposing their shared-residual dependency, this work turns them into one ordered rule: identify reusable computation first, then route what remains. \method{} realizes this rule through token-adaptive shared width, shared content, and residual expert count. Across vision and language tasks, the resulting allocation improves predictive performance while reducing activated computation and measured inference overhead. Ablations and routing analyses confirm that the three stages are complementary, establishing shared-residual decomposition as a practical basis for jointly controlling intra-expert and inter-expert computation.

\section*{Acknowledgment}

This work was funded in part by the National Natural Science Foundation of China grant under number 62536004, 62222603, in part by the Key-Area Research and Development Program of Guangdong Province under number 2023B0303030001, in part by the Program for Guangdong Introducing Innovative and Entrepreneurial Teams (2019ZT08X214), and in part by the Science and Technology Program of Guangzhou under number 2024A04J6310, and in part by the Fundamental Research Funds for the Central Universities 2025ZYGXZR021.

\bibliographystyle{IEEEtran}
\bibliography{references}

\clearpage
\appendices
\section{Additional Experimental Details}
\label{sec:appendix-experimental-details}

\subsection{Evaluation Protocol and Statistical Reporting}

\paragraph{DomainBed.} We follow the train-validation selection criterion. Each domain, referred to as an environment in DomainBed, is split into an 80\% in-split and a 20\% out-split. For each held-out test environment, the in-splits of the remaining environments are used for training, and the checkpoint with the highest mean out-split accuracy across these training environments is selected. Its accuracy on the held-out environment's in-split is reported as the test result. We repeat this procedure with every environment held out, average the resulting environment accuracies to obtain one dataset score, and report the mean and standard deviation over three independently seeded runs. The run seed controls the data split and the Python, NumPy, and PyTorch random generators; deterministic cuDNN execution is enabled.

\paragraph{GLUE.} We use the official training and validation splits and select the best epoch according to the validation score. CoLA is evaluated by Matthews correlation, MRPC by the mean of accuracy and F1, and QNLI, MNLI, and RTE by accuracy; MNLI uses the matched validation split. For each task and seed, the learning rate is selected from $\{2\times10^{-5},3\times10^{-5},5\times10^{-5}\}$ using validation performance. We report the mean and standard deviation over three independently seeded runs. The same seed is passed to the Python, NumPy, and PyTorch generators through Hugging Face Accelerate.

\paragraph{Performance Stability.} Tables~\ref{tab:appendix-domainbed-std} and~\ref{tab:appendix-glue-std} expose the variation terms omitted from the compact main tables. For \method{}, each entry is the mean $\pm$ standard deviation over three runs. Baseline variation is retained when it is available from the cited or reproduced result; an entry without a $\pm$ term was reported without a standard deviation, and ``--'' denotes an unavailable result. These variation estimates describe run-to-run stability.

\newcommand{\appms}[2]{$#1\mathbin{\pm}#2$}

\begin{table*}[htbp]
\centering
\caption{Out-of-domain accuracy (\%) with the variation available for the DomainBed main results.}
\label{tab:appendix-domainbed-std}
\begin{tabular}{@{}lccccc@{}}
\toprule
Method & PACS & VLCS & OfficeHome & TerraInc. & DomainNet \\
\midrule
DeiT-S/16 & \appms{86.2}{0.1} & \appms{79.7}{0.0} & \appms{72.2}{0.4} & \appms{42.0}{0.8} & \appms{47.3}{0.2} \\
GMoE & \appms{88.1}{0.1} & \appms{80.2}{0.2} & \appms{\mathbf{74.2}}{\mathbf{0.4}} & \appms{48.5}{0.4} & \appms{48.7}{0.2} \\
EMoE & \appms{87.8}{0.2} & \appms{79.5}{0.4} & \appms{73.1}{0.2} & \appms{45.9}{0.3} & -- \\
EMoE-L & \appms{87.2}{0.4} & \appms{79.6}{0.2} & \appms{72.5}{0.2} & \appms{46.1}{0.4} & -- \\
LFME & \appms{88.7}{0.2} & \appms{79.7}{0.1} & \appms{73.1}{0.2} & \appms{\mathbf{53.4}}{\mathbf{0.4}} & \appms{47.5}{0.0} \\
DMDA & 88.1 & 79.4 & 69.4 & 49.5 & 45.8 \\
PC-MoE & 89.4 & 80.0 & 74.1 & 49.9 & 48.5 \\
DynMoE & 88.4 & 80.3 & 73.6 & \appms{48.9}{0.5} & 48.2 \\
MASS & 88.7 & 81.1 & 73.8 & 47.5 & -- \\
\textbf{\method{}} & \appms{\mathbf{89.6}}{\mathbf{0.1}} & \appms{\mathbf{81.7}}{\mathbf{0.3}} & \appms{\mathbf{74.2}}{\mathbf{0.2}} & \appms{52.6}{0.3} & \appms{\mathbf{49.4}}{\mathbf{0.1}} \\
\bottomrule
\end{tabular}
\end{table*}

\begin{table*}[htbp]
\centering
\caption{GLUE task scores (\%) with the standard deviations omitted from the compact main table.}
\label{tab:appendix-glue-std}
\begin{tabular}{@{}lccccc@{}}
\toprule
Method & CoLA & MRPC & QNLI & MNLI & RTE \\
\midrule
BERT-large & \appms{64.03}{0.54} & \appms{89.36}{0.09} & \appms{92.46}{0.09} & \appms{86.61}{0.26} & \appms{74.37}{0.78} \\
Standard MoE, top-1 & \appms{63.63}{0.20} & \appms{89.81}{0.30} & \appms{92.39}{0.21} & \appms{86.63}{0.17} & \appms{74.01}{0.29} \\
Standard MoE, top-2 & \appms{64.71}{1.21} & \appms{90.18}{1.33} & \appms{92.53}{0.07} & \appms{86.73}{0.43} & \appms{72.32}{3.54} \\
Standard MoE, top-4 & \appms{64.12}{1.42} & \appms{89.74}{0.40} & \appms{92.65}{0.09} & \appms{86.59}{0.16} & \appms{75.33}{0.95} \\
Standard MoE, top-8 & \appms{64.37}{1.14} & \appms{90.35}{0.68} & \appms{92.49}{0.11} & \appms{86.51}{0.20} & \appms{73.53}{2.21} \\
Standard MoE, average & \appms{64.21}{0.45} & \appms{90.02}{0.29} & \appms{92.52}{0.11} & \appms{86.62}{0.09} & \appms{73.80}{1.24} \\
Standard MoE, best & \appms{64.71}{1.21} & \appms{90.35}{0.68} & \appms{92.65}{0.09} & \appms{86.73}{0.43} & \appms{75.33}{0.95} \\
DynMoE & \appms{65.17}{0.26} & \appms{90.64}{0.26} & \appms{92.59}{0.08} & \appms{86.37}{0.13} & \appms{73.41}{1.96} \\
MASS & \appms{65.62}{1.25} & \appms{90.58}{0.45} & \appms{92.62}{0.13} & \appms{86.74}{0.07} & \appms{75.33}{1.12} \\
\textbf{\method{}} & \appms{\mathbf{66.83}}{\mathbf{0.19}} & \appms{\mathbf{91.57}}{\mathbf{0.26}} & \appms{\mathbf{93.10}}{\mathbf{0.12}} & \appms{\mathbf{86.84}}{\mathbf{0.07}} & \appms{\mathbf{75.47}}{\mathbf{0.91}} \\
\bottomrule
\end{tabular}
\end{table*}

\newpage

\subsection{Architecture and Optimization Details}

\paragraph{Vision configuration.} The vision backbone is ImageNet-pretrained DeiT-S/16 with model dimension $d=384$ and FFN width $H=1536$. Transformer layers 8 and 10 under zero-based indexing are converted to \method{} layers. Each converted layer contains one shared expert, $K=6$ residual experts, and $B=8$ blocks per expert. We use hidden dropout 0.1, stochastic-depth rate 0.1. Images are normalized with ImageNet statistics. Training-domain images use random resized cropping to $224\times224$, horizontal flipping, color jitter, and random grayscale; validation and test images are resized directly to $224\times224$. Table \ref{tab:appendix-domainbed-hparams} lists the final DomainBed optimization settings.

\begin{table}[htbp]
\centering
\caption{Final DomainBed optimization settings. All runs use Adam, a batch size of 32 per training environment, and an evaluation batch size of 64.}
\label{tab:appendix-domainbed-hparams}
\begin{tabular}{@{}lccr@{}}
\toprule
Dataset & Learning rate & Weight decay & Steps \\
\midrule
PACS & $3\times10^{-5}$ & $10^{-6}$ & 5,001 \\
VLCS & $3\times10^{-5}$ & $10^{-6}$ & 5,001 \\
OfficeHome & $1\times10^{-5}$ & $10^{-6}$ & 5,001 \\
TerraInc. & $5\times10^{-5}$ & $10^{-4}$ & 5,001 \\
DomainNet & $5\times10^{-5}$ & $0$ & 15,001 \\
\bottomrule
\end{tabular}
\end{table}

\paragraph{Held-out environments.} PACS contains Art, Cartoon, Photo, and Sketch; VLCS contains Caltech101, LabelMe, SUN09, and VOC2007; OfficeHome contains Art, Clipart, Product, and Real World; TerraIncognita contains locations L100, L38, L43, and L46; and DomainNet contains Clipart, Infograph, Painting, Quickdraw, Real, and Sketch. Each domain is used once as the held-out test environment.

\paragraph{Language configuration.} The language backbone is BERT-large-cased. Transformer layers 20 and 22 under zero-based indexing are converted to \method{} layers with $K=16$ residual experts and $B=16$ blocks. We train for at most 10 epochs with FP16 mixed precision, dynamic padding, maximum sequence length 128, training and evaluation batch sizes of 32, and no gradient accumulation. Optimization uses AdamW with zero weight decay, a linear learning-rate schedule, and no warm-up. The best validation epoch is retained, and training stops after four consecutive epochs without improvement.

\subsection{Computing Environment}

Experiments are run on Ubuntu 20.04.2 LTS workers equipped with an AMD EPYC 75F3 CPU, 503 GiB of host memory, and an NVIDIA GeForce RTX 3090 GPU. The software environment uses Python 3.8.20, PyTorch 2.4.1 with CUDA 12.1 and cuDNN 9.1, torchvision 0.19.1, timm 1.0.20, Tutel 0.2, Transformers 4.46.3, Datasets 3.1.0, Evaluate 0.4.6, and Accelerate 1.0.1.

\section{Generative AI Use}

Generative AI tools were used solely for language editing and polishing. The authors reviewed and verified all resulting text and take full responsibility for the content of the paper.

\end{document}